# GOSt-MT: A Knowledge Graph for Occupation-related Gender Biases in Machine Translation

Orfeas Menis Mastromichalakis[1,*], Giorgos Filandrianos[1], Eva Tsouparopoulou[2], Dimitris Parsanoglou[3], Maria Symeonaki[2] and Giorgos Stamou[1]

[1]*Artificial Intelligence and Learning Systems Laboratory, National Technical University of Athens, Greece*
[2]*Department of Social Policy, Panteion University of Social and Political Sciences, Athens, Greece*
[3]*Department of Sociology, National and Kapodistrian University of Athens, Greece*

**Abstract**
Gender bias in machine translation (MT) systems poses significant challenges that often result in the reinforcement of harmful stereotypes. Especially in the labour domain where frequently occupations are inaccurately associated with specific genders, such biases perpetuate traditional gender stereotypes with a significant impact on society. Addressing these issues is crucial for ensuring equitable and accurate MT systems. This paper introduces a novel approach to studying occupation-related gender bias through the creation of the GOSt-MT (Gender and Occupation Statistics for Machine Translation) Knowledge Graph. GOSt-MT integrates comprehensive gender statistics from real-world labour data and textual corpora used in MT training. This Knowledge Graph allows for a detailed analysis of gender bias across English, French, and Greek, facilitating the identification of persistent stereotypes and areas requiring intervention. By providing a structured framework for understanding how occupations are gendered in both labour markets and MT systems, GOSt-MT contributes to efforts aimed at making MT systems more equitable and reducing gender biases in automated translations.

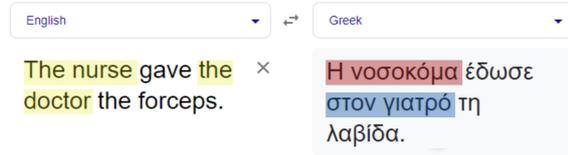

**Figure 1:** An example of gender stereotypes reflected in a translation from Google Translate

## 1. Introduction

Gender bias in machine translation systems is a pervasive issue that compromises the accuracy and fairness of automated translations. Such biases can reinforce harmful stereotypes and contribute to gender inequality, particularly in the context of occupational terms. This problem is exacerbated when MT systems, widely used in diverse applications, systematically associate certain professions with specific genders. Consider the example of Figure 1 where "the doctor" without any gender indication, is translated into Greek as "ο γιατρός" (the male doctor), while "the nurse" is consistently rendered as "η νοσοκόμα" (the female nurse). This illustrates how MT systems can reinforce gender stereotypes by associating certain occupations predominantly with one gender. Such biases are not only misleading but also detrimental, as they perpetuate traditional gender roles and contribute to the gender disparities observed in various professional sectors. Addressing and mitigating these biases is critical to ensure that technology promotes gender equality rather than perpetuating discrimination.

Our motivation derives from two primary concerns: the persistent gender inequalities in the labour market and the existence of gendered algorithmic bias [1, 2, 3, 4], as they are both highlighted in strategic social policy documents such as the European Commission's Gender Equality Strategy 2020-2025 (EU Commission, 2020) [1]. The Commission emphasises the necessity of challenging gender stereotypes, which are fundamental drivers of gender inequality across all societal domains, and identifies gender stereotypes as significant contributors to the gender pay and pension gaps. Moreover, the Strategy places a specific focus on the impact of Artificial Intelligence, highlighting the need for further exploration of its potential to amplify or contribute to gender biases. Specifically, gender bias in machine translation systems is a significant element of this aspect.

For identifying bias and its source, it is essential to incorporate external knowledge that accurately reflects the actual world, such as the distribution of occupations in actual labour markets and within training datasets among different countries and languages. This underscores the importance of employing tools like Knowledge Graphs to refine and improve AI systems, ensuring they support fairness and transparency in decision-making.

Semantic information and specifically Knowledge Graphs (KGs) have become increasingly prominent as tools that enhance machine learning systems, particularly in areas like explainable AI (XAI), [5, 6, 7, 8, 9, 10], fairness, fact checking [11, 12, 13], and reasoning [14, 15, 16, 17]. They serve as foundational elements by structuring vast datasets, which grounds large language models and other AI technologies with a well-organized layer of knowledge [18]. This structured knowledge is essential for addressing critical issues, such as gender bias, ensuring these systems operate within ethical guidelines [18, 19].

Our research aims to investigate both horizontal and vertical occupational gender segregation, and how these phenomena manifest in various types of gender bias in machine translation, in English, French, and lower-resource languages like Greek. In this paper, we propose a novel approach to studying occupation-related gender bias in MT systems through the creation of a Knowledge Graph (KG) on

---

---

[1]https://commission.europa.eu/strategy-and-policy/policies/justice-and-fundamental-rights/gender-equality/gender-equality-strategy_en

Gender and Occupation Statistics for Machine Translation (GOSt-MT). Built upon the International Standard Classification of Occupations (ISCO-08), a hierarchical framework endorsed by the International Labour Organisation (ILO, 2012) that categorizes occupations into groups on different levels, GOSt-MT incorporates real labour statistical data and statistical data from textual corpora to support and facilitate the detection and study of stereotypical automatic translations. By integrating structured occupational classifications and comprehensive gender statistics into a Knowledge Graph, we offer a nuanced understanding of how occupations are "gendered" in both actual labour markets and MT training datasets, offering insights into discriminating and resisting gender biases in the world(s) of employment with twofold utility: identifying recurring stereotypical representations that resist despite reality, i.e. existing data, being fundamentally different; and mapping professional areas that still require interventions to overcome gender imbalances. This work contributes to the broader effort of making MT systems more equitable and reliable, promoting gender fairness and eliminating stereotypes in various societal domains.

## 2. Related Work

Recent research has increasingly focused on uncovering and mitigating gender biases in machine translation systems. Notably, studies such as those by [20] have empirically demonstrated how commercial translation systems often perpetuate gender stereotypes by assigning genders to professions based on societal biases rather than linguistic accuracy. Similarly, [21] highlighted the tendency of translation algorithms to prefer masculine pronouns even in contexts where gender is unspecified. In the era of Large Language Models, the study [22] reveals that tools like ChatGPT[2], Google Translate[3], and Microsoft Translator[4] perpetuate gender defaults and stereotypes, particularly failing to translate the English gender-neutral pronoun "they" into equivalent gender-neutral pronouns in other languages, resulting in translations that are incoherent and incorrect, especially for low-resource languages. This conclusion also holds for high-resource languages such as Italian, as the preliminary analysis in [1] demonstrates that ChatGPT's performance across different scenarios reveals a strong male bias, particularly when not explicitly prompted to consider gender alternatives.

Furthermore, studies such as [20, 23, 24, 25] have highlighted how gender biases manifest in the assignment of pronouns to professions in machine translation systems. Professions like doctors, engineers, and presidents are frequently associated with male pronouns, while roles like dancers, nurses, and teachers are typically linked to female pronouns. Moreover, language models have been shown to override explicit gender information in translations; for example, a translation from English to Spanish incorrectly changed the gender of a female doctor to male, as noted in [25].

This leads to a systematic failure to include feminine and gender-neutral options, underscoring the need for ongoing improvements in machine translation models to ensure they align with evolving societal norms and support inclusive communication.

Knowledge Graphs (KGs) have been increasingly utilized to promote responsible and fair AI applications [26]. For instance, [27] provides a comprehensive survey on bias in AI and highlights the role of KGs in detecting and correcting biases, demonstrating how integrating KGs with machine learning models can enhance the transparency and accountability of AI applications. To the best of our knowledge, no existing works have utilized statistics from knowledge graphs to identify biases in MT systems. This research aims to fill this gap by providing a valuable resource to the community, specifically for identifying occupational gender biases and tracing their origins for machine translation systems.

## 3. Methodology

In this section, we delve into the methods and techniques employed to create the GOSt-MT Knowledge Graph. GOSt-MT integrates statistics from both real-world labour data and textual datasets, providing a comprehensive resource for analyzing gender bias in machine translation. To achieve this, we utilized multiple sources, including national and European statistical agencies and databases, to extract accurate and up-to-date labour statistics. In addition, we developed a pipeline to extract gender statistics from textual datasets, ensuring thorough analysis and integration of diverse data sources. The following subsections detail our methodology and the specific resources and tools we employed throughout this work, highlighting the steps taken to ensure the accuracy and reliability of the GOSt-MT Knowledge Graph.

### 3.1. Real World Statistics

For the purpose of this study, we conducted secondary analyses using mainly data from EUROSTAT's labour market participation indicators, which are drawn from the European Union Labour Force Survey (EU-LFS), (EUROSTAT 2022, 2024), the National Statistical Authorities of Greece (ELSTAT[5]) and the UK's NOMIS-Office for National Statistics (ONS)[6]. The EU-LFS is a European large scale sample survey that provides quarterly and annual statistics on labour market participation and inactivity among individuals aged 15 and older. It is the largest survey of its kind in Europe, offering extensive data that ensures comparability across countries and over time due to its standardised definitions, classifications, and variables. The survey follows guidelines set by the International Labour Organisation (ILO) and employs standard classifications such as the International Standard Classification of Occupations (ISCO-08), detailing occupations at the 4-digit level for the current main job and at the 3-digit level for the last job. Publicly available statistics include employment data by detailed occupation (ISCO-08 2-digit level), broken down by age and gender. More specifically, we employed secondary statistical analysis on EUROSTAT's and ELSTAT's data to estimate the gendered distributions of occupations in Greece (2011-2022 at ISCO-08 3-digit level), the UK (2013-2019) at ISCO-08 2-digit level and France (2013-2022) at the same level. Analysis was also performed on NOMIS-ONS data for the UK 2020-2023 to produce results for the gender distribution of occupations at SOC2020 4-digit level. SOC2020 stands for the Standard Occupational Classification 2020, a system used in the UK to classify and categorise occupations. SOC2020 is developed by ONS and is used for a variety of purposes, including statistical analyses and labour market studies. SOC2020 is based upon the same classification principles as the 2008 version of its international equivalent ISCO-08.

---

[2]https://chatgpt.com/
[3]https://translate.google.com
[4]https://translator.microsoft.com

[5]https://www.statistics.gr/en/home/
[6]OfficialCensusandLabourMarketStatistics,ONS,https://www.nomisweb.co.uk/datasets/aps218/reports/employment-by-occupation?compare=K02000001

More specifically, the occupational distributions were estimated based on the availability of data in each country, i.e. the gendered distribution at the 2-digit level was calculated for all three countries, encompassing 43 respective occupations for males and females from 2011-2022 for Greece, 2013-2022 for France, and 2013-2019 for the UK. For Greece, gendered distributions at the 3-digit level were also estimated based on secondary data analysis from the National Statistical Authority of Greece and the Mechanism of Labour Market Diagnosis provided by the Hellenic Republic, Ministry of Labour and Social Insurance. This analysis provides gendered distributions for 130 occupations at that level from 2011-2022. Additionally, for the UK, based on data from NOMIS-ONS, distributions for 412 occupations are estimated for the years 2020-2023, encompassing the period following Brexit. Moreover, a secondary analysis was conducted for specific 3-d level occupations, such as doctors, which were not publicly available from EUROSTAT for France. This examination utilised data from the OECD Data Explorer archive[7] and the World Health Organization's European Health Information Gateway[8].

Calculating the respective percentages for the three countries from 2011 and onwards enabled an examination of the evolution of these distributions over time, revealing the occupational gender segregation trends over these periods in the specified countries. For instance, Figure 2 illustrates the changes in gender distribution among medical practitioners (doctors) over the past decade. The statistics depicted in this Figure reveal notable trends and differences across Greece, France, and the UK. In Greece, male doctors ranged from 56.53% to 63.88%, with a significant decline to 56.53% in 2022, while female doctors increased from 36.12% to 43.47%, indicating a shift towards gender balance. The UK shows a more balanced distribution, with male doctors decreasing from 55.43% in 2011 to 50.56% in 2023, and female doctors rising from 44.57% to 49.44%. France also trends towards gender balance, with male doctors decreasing from 60.61% in 2011 to 52.52% in 2021, and female doctors increasing from 39.39% to 47.48%. Comparatively, the UK has the most stable gender distribution, approaching parity by 2023, while France follows closely behind. Greece, although showing improvement, still has a more pronounced gender disparity. This analysis highlights the dynamic nature of gender distribution in the medical profession and the varying rates of progress across these countries.

Further analysis on the gender distribution in occupations across Greece, France, and the UK reveals consistent patterns of gender disparity. Technical and manual trades are predominantly male in all three countries, indicating a significant gender imbalance in sectors requiring technical skills and manual labour. Conversely, care-giving and administrative roles are predominantly female, reflecting societal trends where women are more represented in these fields. However, the medical profession should not be considered predominantly male, as evidenced by the nearly equal percentages in the UK (50.56% male and 49.44% female) and the substantial female representation in Greece (57.53% male and 43.47% female for 2023) according to the latest available data. Conversely, midwifery nurses can be categorically considered female, as the respective percentage is equal to 100%.

---

[7]https://data-explorer.oecd.org/
[8]https://gateway.euro.who.int/en/

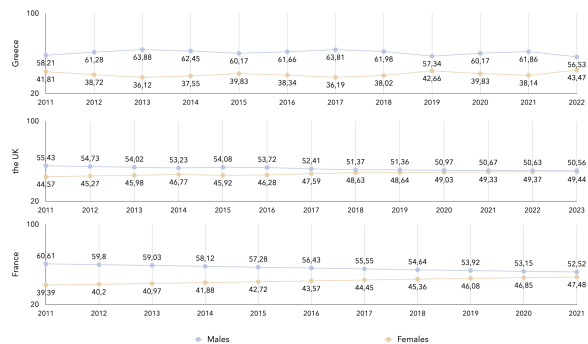

**Figure 2:** Gender distribution of medical practitioners in Greece (2011-2022), the UK (2011-2023), and France (2011-2021)

### 3.2. Dataset Statistics

Our methodology for extracting gender statistics for occupations from textual datasets involves a comprehensive three-fold pipeline. As illustrated in Figure 3, this pipeline consists of three sequential modules designed to detect, link, and analyze occupational terms and their associated genders within textual data. The first module focuses on detecting occupations within a given text. Using a Large Language Model (LLM), this module scans the text to identify and extract occupational terms accurately. Once the occupations are detected, the second module comes into play, linking these terms to the corresponding occupations in the GOSt-MT Knowledge Graph. This linking process ensures that each detected occupation is mapped to a standardized occupational classification, facilitating consistent analysis and comparison. The third module is dedicated to identifying the gender associated with each detected occupation. This module determines the gender references within the context of the text, enabling us to compile precise gender statistics for each occupation. Through this pipeline, we are able to generate detailed statistics on the gender distribution of occupations within a textual dataset. These statistics are then incorporated into the GOSt-MT, enriching the Knowledge Graph with valuable data on gender representation.

#### 3.2.1. Occupation Extraction

For the Occupation Extraction module, we employed a Large Language Model (LLM) to detect occupations in a given text. The LLM was instructed in a zero-shot prompting manner to identify occupations from the text, along with their respective contextual references and corresponding descriptions. The latter facilitated matching the identified occupations with the corresponding occupations in the GOSt-MT KG (see 3.2.2 for further details) as well as mitigating LLM hallucinations. As an illustration, consider the following example of an input sentence and the respective output of the Occupation Extraction module (this example was created with

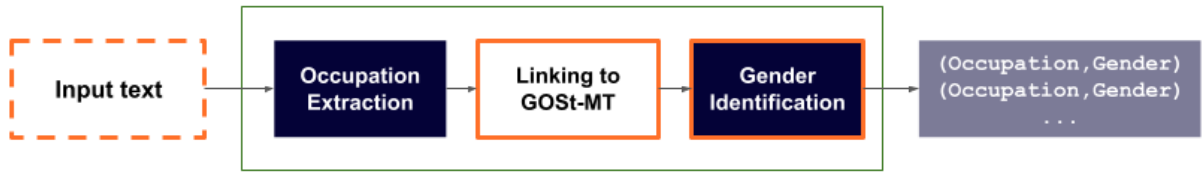

**Figure 3:** The GOSt-MT pipeline for calculating dataset statistics

the Llama-70b [9] [28] model as our deployed LLM):

**Example 1.** Detecting occupations in text
**Input:**
*The doctor put the cast on my leg while talking to the nurses about his new car.*

**Output:**
*Occupation title: Doctor
Appearing in text as: doctor
Description: A medical professional who diagnoses and treats illnesses and injuries.*

*Occupation title: Nurse
Appearing in text as: nurses
Description: A healthcare professional who assists doctors and provides hands-on care to patients*

We experimented with multiple LLMs including variations of LLama2 [28], Mistral, Mixtral [29], Tower [30], and Meltemi [31]. The results across the models were very similar, due to the simplicity of the task, particularly in cases where one or more occupations were referred to in the texts. For the final results, we utilized Mixtral-8x70-v0.1 [10], which empirically has shown the best performance [11].

While experimenting with the LLMs we identified two primary forms of hallucinations and addressed them separately. The first form involves the LLM detecting occupations that are not present in the text. To address this, we asked the LLM to provide the in-text form of the detected occupations along with their titles and descriptions. We then used fuzzy string matching to verify that these detected terms were indeed part of the input text. If a detected term did not match any words in the input text above a certain threshold, it was disregarded as a hallucination. The second form of hallucination occurs when the LLM incorrectly identifies non-occupational terms as occupations. This issue was particularly prevalent with smaller models and in cases where no occupations were present in the input text. We addressed this form of hallucination using the second module of our pipeline, which is described in detail in the following subsection.

### 3.2.2. Linking to GOSt-MT

To ensure the occupations detected by the Large Language Model in the first stage align with the GOSt-MT Knowledge Graph that is curated by domain specialists, we implemented a linking module. Since the KG is based on ISCO-08, which includes not only an occupation taxonomy but also descriptions for each occupation, we framed this task as a retrieval problem. The descriptions generated by the LLM for each detected job title are used to retrieve the most closely matching occupation from the KG. To accomplish this, we converted both the descriptions of each occupation in the KG and those generated by the LLM into embeddings. Following the approach proposed by [32], we utilized angle-based embeddings to map the descriptions into a latent space where they can be easily compared. We then used cosine similarity as the distance metric to find the closest matching descriptions. In the following example, you can see the occupations of the GOSt-MT KG that matched the detected occupations of the previous step illustrated in Example 1.

By setting a similarity threshold, we can effectively filter out hallucinations where a detected term is misidentified as an occupation. If the similarity between the LLM-provided description and any existing occupation description in the KG falls below this threshold, the detected occupation is disregarded. This retrieval and embedding-similarity approach helps us ensure that only valid occupations, as defined in our curated KG, are considered, thereby addressing potential hallucinations from the initial detection stage. By rigorously matching descriptions, we maintain the accuracy and reliability of the occupation data integrated into the GOSt-MT Knowledge Graph.

### 3.2.3. Gender Identification

The final and most challenging part of our pipeline is the gender identification module. This module aims to identify the gender of an occupation in the text or conclude that the gender cannot be determined from the context. By doing this, we can calculate gender statistics for the occupations detected and matched with GOSt-MT in the previous stages and ultimately incorporate these statistics into the Knowledge Graph. We identified three distinct cases for deriving the gender of an occupation, which we investigate stepwise.

**Example 2.** Linking the detected occupations to GOSt-MT

**Doctor → Medical Doctor** *(ISCO code: 221): Medical doctors (physicians) study, diagnose, treat and prevent illness, disease, injury, and other physical and mental impairments in humans through the application of the principles and procedures of modern medicine. They plan, supervise and evaluate the implementation of care and treatment plans by other health care providers, and conduct medical education and research activities.*
**Nurse → Nursing and midwifery professional** *(ISCO code: 222): Nursing and midwifery professionals provide treatment and care services for people who are physically or mentally ill, disabled or infirm, and others in need of care due to potential risks to health including before, during and after childbirth. They assume responsibility for the planning, management and evaluation of the care of patients, including the supervision of other health care workers, working*

---

*autonomously or in teams with medical doctors and others in the practical application of preventive and curative measures.*

If one case determines the gender, we do not proceed to the next steps. The first case occurs when the occupation word itself indicates gender. This is common in notional gender languages such as English as well as grammatical gender languages such as Spanish, French, and Greek, where variations in words often signify gender (e.g. waiter/waitress, or in Greek "νοσοκόμος" for a male nurse and "νοσοκόμα" for a female nurse). We use the SpaCy[12] library to automatically detect if a word has a gender indication. If the occupation word does not indicate gender, we proceed to the second case, where gender is directly mentioned through pronouns. For example, in the sentence "He is a nurse", the pronoun "He" directly indicates the gender of the nurse. To identify such cases, we construct the syntactic dependency tree using SpaCy and check for any direct links from a gendered pronoun to the occupation. If neither the occupation word nor direct pronouns indicate gender, we move to the third case: gender indication through coreference. Consider the text, "Today the doctor came to the hospital 45 minutes late. Consequently, his first appointment had already left." Here, the gender of "doctor" is inferred from the pronoun "his" in the second sentence. For this, we use the Coreferee[13] library to find all linguistic expressions (also called mentions) in the given text that refer to the same entity, here the occupation of interest. We then check the gender of the words and pronouns linked to the occupation. If we find a gender indication, we determine the occupation's gender; if not, we conclude that the gender cannot be determined from the text and exclude this detection from our statistics. Consider the example below that follows Examples 1, and 2 and illustrates the output of the Gender Identification module for the input of Example 1.

**Example 3.** Identifying the gender of the detected occupations
**Doctor → Male** *(Coreference)*
**Nurse → Not Clear*

## 4. The Knowledge Graph

Based on the methodology described in Section 3, we collected real-world gender statistics on the labour market as well as occupation-related gender statistics from textual datasets. In this work, we have focused on employment data from the UK, Greece, and France, and we have extracted the respective statistics for English, Greek, and French, from the WMT dataset [14] as well as a part of the C4 dataset [33] [15]. This extensive data collection enabled us to create the GOSt-MT Knowledge Graph. By systematically integrating structured occupational classifications with comprehensive gender statistics, we have constructed a detailed and accurate representation of gender distribution across various occupations.

The GOSt-MT Knowledge Graph serves as a resource for studying gender bias in machine translation systems and providing valuable insights into gender representation within different professional sectors. This comprehensive approach allows for a nuanced understanding of how occupations are "gendered" in both the actual labour market and the textual data used to train MT systems.

For example, by analyzing the WMT dataset, a widely used resource for training machine translation systems, we discovered a consistent gender misalignment in the occupational category of lawyers. Specifically, in over 85% of instances where a gender was assigned to a lawyer, it was male rather than female. This bias could potentially be transferred to a machine translation model trained using this dataset, perpetuating a stereotype with significant societal impact. Moreover, such biases are not only detrimental to societal equality but also fail to accurately represent the distribution of this profession in the real world. Specifically, real-world statistics from 2011 to 2022 show that each year there were consistently more female lawyers, with the ratio ranging from 56% to 62%.

This analysis underscores that, beyond computer scientists and AI researchers, the GOSt-MT is also of great interest to social researchers and scholars in fields such as Science and Technology Studies (STS). It provides a robust tool for examining the intersection of technology and gender, offering a valuable resource for those aiming to address and mitigate gender biases in both technology and society.

### 4.1. Structure

The structure of the GOSt-MT Knowledge Graph is presented in Figure 4. GOSt-MT is fundamentally based on the International Standard Classification of Occupations (ISCO-08), which provides a hierarchical taxonomy of occupations. This hierarchy organizes occupations in our KG into broader and narrower categories linked through "subclassOf" relations. For example, "Professionals" (ISCO Code 2) includes "Health Professionals" (ISCO Code 22) as a subclass, which further branches into several occupations including "Medical Doctors" (ISCO Code 221) and "Nursing and Midwifery Professionals" (ISCO Code 222). Each occupation in the KG has a title, description, and ISCO code, all extracted from the ISCO-08 standard.

The GOSt-MT Knowledge Graph also integrates comprehensive statistical data about gender representation in various occupations. This integration is achieved through "Statistics" entities, which link occupations to gender statistics. Each "Statistics" entity includes two key attributes: malePercentage and femalePercentage. These percentages indicate the proportion of male and female workers in a given occupation, or the respective proportion of masculine and feminine mentions of occupations in textual corpora.

The "Statistics" entities are connected to either a "Dataset" entity or a "Survey" entity, depending on the source of the data. Each "Dataset" entity includes a title and description, reflecting the dataset's content. If the statistics are derived from a survey, the "Survey" entity also includes a title, description, and the year or time period of the survey.

Furthermore, each "Statistics" entity is linked to a "Country" entity, providing contextual information about the geographical origin of the data or the language of the textual corpora respectively. When the statistics are linked to a dataset, the relationship is represented by the "hasLanguage" relation, indicating the language of the analyzed texts. Conversely, if the statistics are from a survey, the "linkedToCountry" relation specifies the country from which the survey data originated and refer to.

---
[12] https://spacy.io/api/morphology
[13] https://pypi.org/project/coreferee/
[14] https://huggingface.co/datasets/wmt/wmt14
[15] https://huggingface.co/datasets/allenai/c4

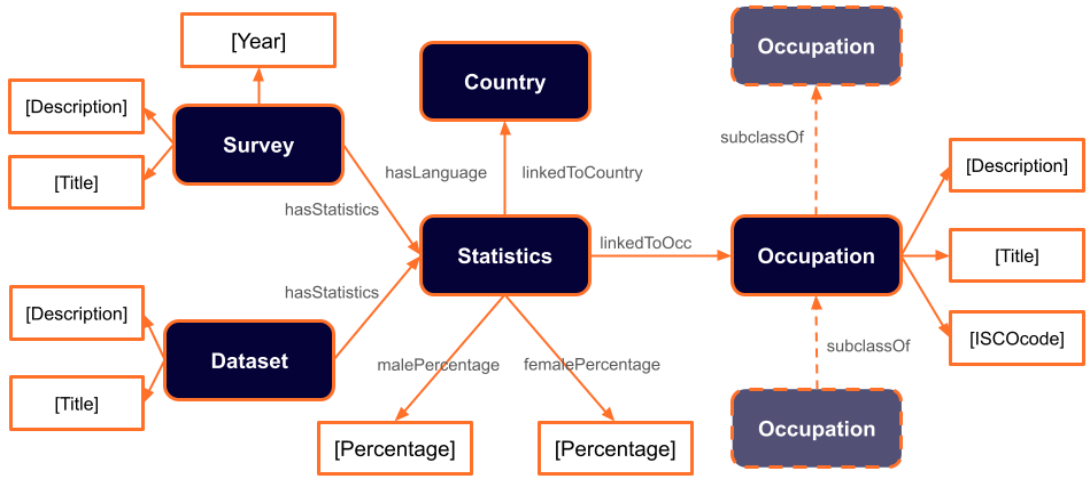

**Figure 4:** The GOSt-MT KG structure

## 5. Conclusion & Future Work

This study highlights the significant challenges posed by gender bias within machine translation (MT) systems, particularly regarding the representation of occupational roles. The development of the GOSt-MT Knowledge Graph represents a novel approach to integrating real-world labour statistics with the textual corpora used in MT training. By combining statistics from multiple sources into a single knowledge graph, we provide an opportunity to study and identify misalignments among the occupational distributions across genders in the real world and the training sets of MT models.

Future work will focus on expanding our methodology to include a broader array of datasets, thereby enriching the statistical analysis available for commonly used training corpora in large language models. Additionally, using the GOSt-MT pipeline to identify occupational titles and their genders will be crucial for detecting discrepancies in gender representation of occupations between the input and output of MT systems. Subsequently, GOSt-MT could be employed to identify the sources of these misalignments, whether they arise from the datasets, inherent algorithmic biases, or a combination of both.

## 6. Limitations

This study, is subject to certain limitations that must be acknowledged. First, the statistics integrated into the GOSt-MT Knowledge Graph are derived exclusively from European and UK labour markets. This regional focus may limit the generalizability of our findings to other geographic areas where occupational roles and gender distributions may differ significantly.

Additionally, the GOSt-MT pipeline itself may not be entirely free from biases, similar to those it aims to identify. A particular point of concern is the coreference model, which relies on a language model potentially vulnerable to the same gender biases we seek to identify. While these models have been specifically trained to mitigate such biases—thereby making them less susceptible—it is crucial to recognize that no model is completely immune to bias. This was a decisive factor in opting for these specialized models over more general large language model (LLM) techniques, which may not have the same focus on minimizing gender bias.

Lastly, the GOSt-MT pipeline's applicability to languages with limited available data represents another limitation. For languages that lack substantial textual or labour market data, the effectiveness of the GOSt-MT in detecting gender biases may be compromised. This underscores the need for future research to adapt and refine the pipeline for broader linguistic coverage, ensuring that the benefits of this research can be extended to a wider array of languages and cultural contexts.

## Acknowledgments

This research work is Co-funded by the European Union's Horizon Europe Research and Innovation programme under Grant Agreement No 101070631 and from the UK Research and Innovation(UKRI) under the UK government's Horizon Europe funding guarantee (Grant No 10039436).-FSTP – Pilot Project:- SURE-GB